\documentclass[sigconf]{acmart}

\copyrightyear{2026}
\acmYear{2026}
\setcopyright{cc}
\setcctype{by}
\usepackage{array}
\usepackage{amsmath}
\usepackage{booktabs}
\usepackage{graphicx}
\usepackage{tabularx}
\usepackage{url}
\usepackage{balance}
\newcolumntype{Y}{>{\raggedright\arraybackslash}X}

\newlength{\EgoLinkAuthorRowSep}
\makeatletter
\patchcmd{\@mkauthors@iii}
  {\lineskip=1pc\relax}
  {\lineskip=\EgoLinkAuthorRowSep\relax}
  {}
  {\PackageWarning{egolink}{Could not adjust the author-row spacing}}
\makeatother

\acmConference[MM '26]{Proceedings of the 34th ACM International Conference on Multimedia}{November 10--14, 2026}{Rio de Janeiro, Brazil}
\acmBooktitle{Proceedings of the 34th ACM International Conference on Multimedia (MM '26), November 10--14, 2026, Rio de Janeiro, Brazil}
\acmDOI{10.1145/3767308.3837706}
\acmISBN{979-8-4007-2213-4/2026/11}

\begin{document}

\title[EgoLink Grand Challenge at ACM Multimedia 2026]{EgoLink: Egocentric Language-Vision Interactive Network Knowledge Challenge Summary}

\author{Renhe Sun}
\authornote{These authors contributed equally to this work.}
\affiliation{\institution{Ant Group}\country{Shanghai, China}}

\author{Yueying Feng}
\authornotemark[1]
\affiliation{\institution{Zhejiang University}\country{Hangzhou, China}}

\author{Bohan Yu}
\authornotemark[1]
\affiliation{\institution{Ant Group}\country{Shanghai, China}}

\author{Tong Niu}
\authornotemark[1]
\affiliation{\institution{Ant Group}\country{Beijing, China}}

\author{Yunqi Liu}
\authornotemark[1]
\affiliation{\institution{Ant Group}\country{Beijing, China}}

\author{Zitong Wang}
\affiliation{\institution{Ant Group}\country{Hangzhou, China}}

\author{Haolin He}
\affiliation{\institution{The Chinese University of Hong Kong}\country{Hong Kong, China}}

\author{Kexin Shi}
\affiliation{\institution{Ant Group}\country{Shanghai, China}}

\author{Hualei Wang}
\affiliation{\institution{Ant Group}\country{Shanghai, China}}

\author{Qiuqiang Kong}
\affiliation{\institution{The Chinese University of Hong Kong}\country{Hong Kong, China}}

\author{Chang Yao}
\affiliation{\institution{Zhejiang University}\country{Hangzhou, China}}

\author{Yuge Huang}
\affiliation{\institution{Ant Group}\country{Shanghai, China}}

\author{Weiqiang Wang}
\affiliation{\institution{Ant Group}\country{Hangzhou, China}}

\author{Jingyuan Chen}
\authornote{Corresponding authors. jingyuanchen@zju.edu.cn rex.lj@antgroup.com}
\affiliation{\institution{Zhejiang University}\country{Hangzhou, China}}

\author{Jian Liu}
\authornotemark[2]
\affiliation{\institution{Ant Group}\country{Hangzhou, China}}

\renewcommand{\shortauthors}{Renhe Sun et al.}

\begin{abstract}
The EgoLink Challenge at ACM Multimedia 2026 benchmarks egocentric multimodal intelligence linking first-person perception, social reasoning, and interactive action. Unlike benchmarks centered on visible objects and activities, EgoLink evaluates whether models can interpret emotional and social cues, resolve visually grounded references, and convert user requests into verified tool-mediated actions.  Its two tracks cover social reasoning in egocentric video and interactive agents that complete service tasks through dialogue, tool use, and closed-loop execution.  Together, they test whether perception, reasoning, and decision-making remain coupled under both offline question answering and state-changing interaction.  We highlight a multi-stage Track 1 data pipeline for multiple-choice question (MCQ) generation, chain-of-thought generation, automatic quality grading, and video-contribution filtering.  We summarize the task design, evaluation metrics, rules, leaderboards, and representative solutions.  The benchmark combines objective answer scoring with process- and result-based evaluation, enabling analysis of both final correctness and execution quality.  The best Track 1 system achieved 0.9263 accuracy, while the best Track 2 system achieved 0.844 joint success and 0.916 micro accuracy.  Participant reports reveal a shared principle: strong systems make intermediate evidence explicit through candidate verification, option-robust reasoning, scenario-aware planning, tool canonicalization, and state-aware correction before answering or acting.  These findings highlight the importance of process verification and evidence-grounded reasoning for reliable egocentric multimodal intelligence.
\end{abstract}



\maketitle

\clearpage

\section{Introduction}

Egocentric multimedia intelligence requires models to interpret the world from the user's situated point of view~\cite{grauman2022ego4d}.  This setting is central to AI glasses, augmented-reality assistants, social robots, and embodied service agents, where a system must reason over camera motion, partial observations, speech, gestures, objects, and ongoing human interactions~\cite{e3dataset,egobench2026}.  In such environments, a correct response often depends on more than recognizing visible entities: the system must identify emotional states, explain social causes, and infer future intents~\cite{e3dataset,kong2025sivbench}, while deciding when a visually grounded request should become an executable action~\cite{egobench2026}.

Existing egocentric benchmarks have advanced first-person perception~\cite{grauman2022ego4d}, whereas EgoLink targets integrated social and interactive intelligence beyond perception.  Two recurring failure modes motivate this focus.  First, social answers can be linguistically plausible but unsupported by the observed video or audio~\cite{guo2023desiq,he2025audio}.  Second, interactive agents can produce fluent final responses while invoking the wrong tool, using an ungrounded entity, or changing the database state before the task preconditions are verified~\cite{yao2024taubench,egobench2026}. Both failures reflect weak evidence--decision coupling.

\begin{figure}[!t]
    \centering
    \includegraphics[width=\columnwidth]{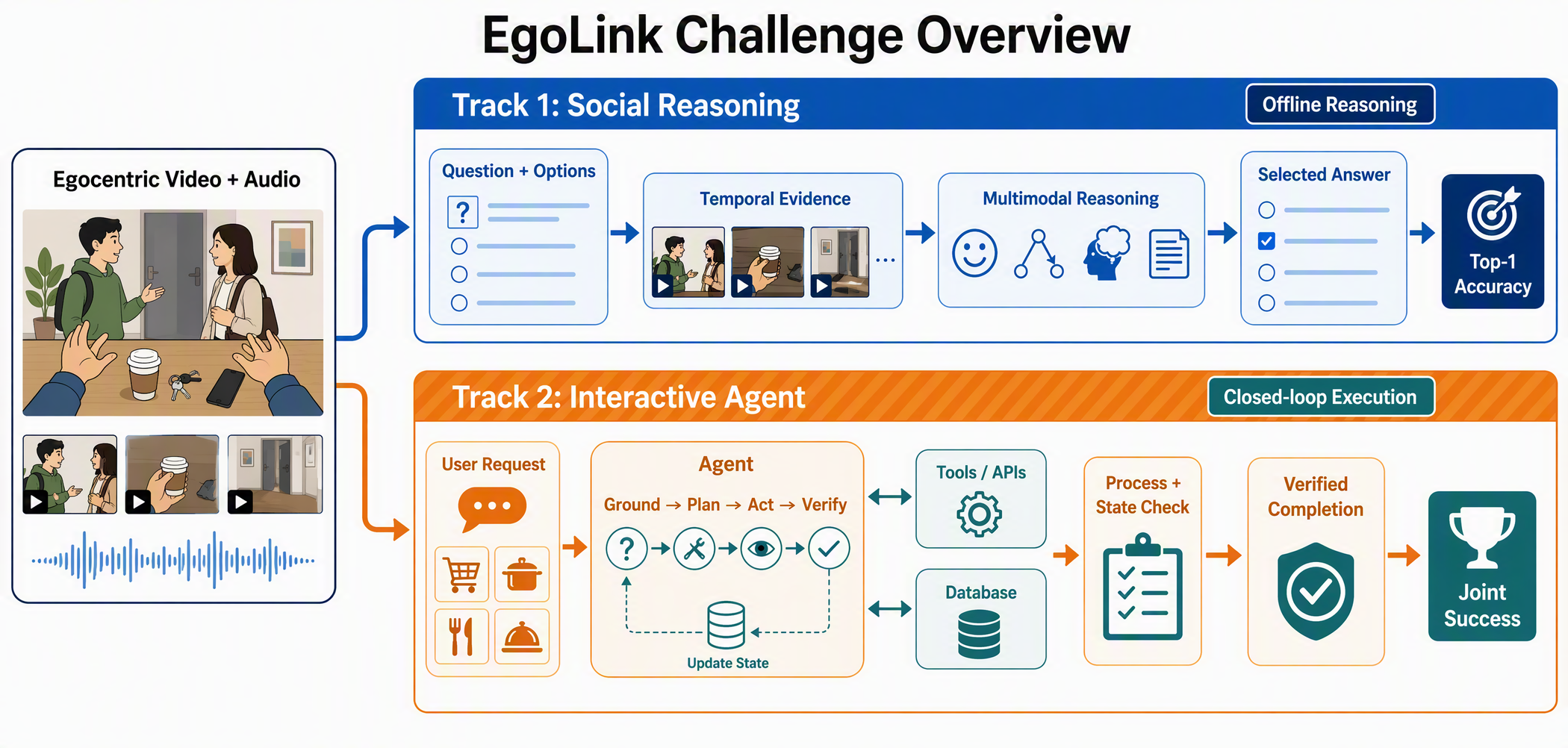}
    \caption{EgoLink challenge overview.}
    \label{fig:overview}
\end{figure}

The EgoLink Challenge\footnotemark{} addresses these issues through two complementary tracks (Figure~\ref{fig:overview}).  Track 1, Social Reasoning in Egocentric Video, uses a unified multiple-choice question (MCQ) protocol to evaluate emotional perception and localization, social causal reasoning, behavioral intent prediction, and egocentric semantic summarization.  Track 2, Interactive Agent Challenge, evaluates whether a service agent can combine first-person visual streams, user instructions, and external tools to complete real-world tasks in retail, kitchen, restaurant, and order scenarios~\cite{egobench2026}.  Together, the tracks test whether perception, reasoning, and decision-making can remain coupled under both offline question answering and closed-loop interaction.

The paper makes three contributions.  First, we present EgoLink as a benchmark that jointly studies egocentric social cognition and interactive task execution.  Second, we detail the Track 1 construction pipeline---question and choice generation, template-guided chain-of-thought generation, five-dimensional quality grading, and video-contribution filtering---together with the official data, rules, evaluation protocol, and leaderboards for both tracks.  Third, we analyze representative submissions and identify a shared emphasis on explicit evidence construction before final prediction or execution.

\section{Related Work}

\subsection{Egocentric Social Reasoning}

Egocentric video understanding has expanded from activity recognition to social and affective interpretation. The $E^3$ dataset provides first-person videos with embodied emotion annotations and forms the basis of EgoLink Track 1~\cite{e3dataset}. EgoLink reformulates these videos into multiple-choice questions for social reasoning. Interpreting gestures, tone, and facial expressions requires temporal and interpersonal context, as similar observations may carry different social meanings~\cite{kong2025sivbench}. DeSIQ further shows that social-interaction MCQs can contain shortcuts exploitable without the intended multimodal context~\cite{guo2023desiq}. These findings motivate EgoLink's emphasis on social answers grounded in observable multimodal evidence.

\footnotetext{\url{https://ego-link.github.io/challenge2026}.}

\subsection{Interactive Multimodal Agents}

Interactive multimodal agents extend visual understanding to tool-mediated task execution. Prior benchmarks study multi-step multimodal planning, visually grounded execution, and stateful tool-agent-user interaction~\cite{ma2024mnms,koh2024visualwebarena,yao2024taubench}. EgoBench formalizes the egocentric setting through a user-agent-tool sandbox with simulated users, official tools, and database-state validation~\cite{egobench2026}. EgoLink Track 2 adapts this setting to interactive service tasks.

Compared with text-only tool-use benchmarks, task entities are often specified visually or deictically. A user may refer to a product or previously observed object without providing its canonical database name. Agents must map these references to valid tool parameters and use tool feedback to guide subsequent actions, with execution evaluated against the resulting database state.

\section{Challenge Description}

\subsection{Tasks}

\subsubsection{Track 1: Social Reasoning in Egocentric Video}
Track 1 evaluates social reasoning in egocentric video through a unified multiple-choice question (MCQ) protocol. It covers four capability groups: emotional perception and temporal localization, social causal reasoning, behavioral intent prediction, and egocentric semantic summarization. Answering requires interpreting temporal, affective, and interpersonal cues beyond object and action recognition.

\subsubsection{Track 2: Interactive Agent Challenge}
Track 2 evaluates interactive service-task execution using first-person visual streams, user instructions, scenario databases, and official tools/APIs. The agent must ground user references, decide whether to answer, request clarification, or invoke tools, and use returned observations to guide subsequent actions. Database updates must be supported by visual evidence and tool observations~\cite{egobench2026}.


\subsection{Datasets and Data Construction}

The Track 1 dataset pairs $E^3$ videos~\cite{e3dataset} with EgoLink-specific labels and MCQs. The release provides training and validation annotations, together with an evaluation benchmark of 4,030 MCQs paired with 1,055 egocentric videos. The challenge annotations must be used in place of the original $E^3$ labels.

MCQs and answer choices are generated from multimodal clips, followed by template-guided chain-of-thought rationales~\cite{wei2022chain}. Quality grading follows the LLM-as-a-judge paradigm~\cite{zheng2023judging}, using Qwen3-235B-A22B~\cite{yang2025qwen3}. Each candidate receives a score on a five-point scale for answer consistency, distractor quality, language fluency, reasoning logic, and simplified reasoning quality. These scores guide quality-based filtering.

\begin{figure}[!t]
    \centering
    \includegraphics[width=\columnwidth,height=0.30\textheight,keepaspectratio]{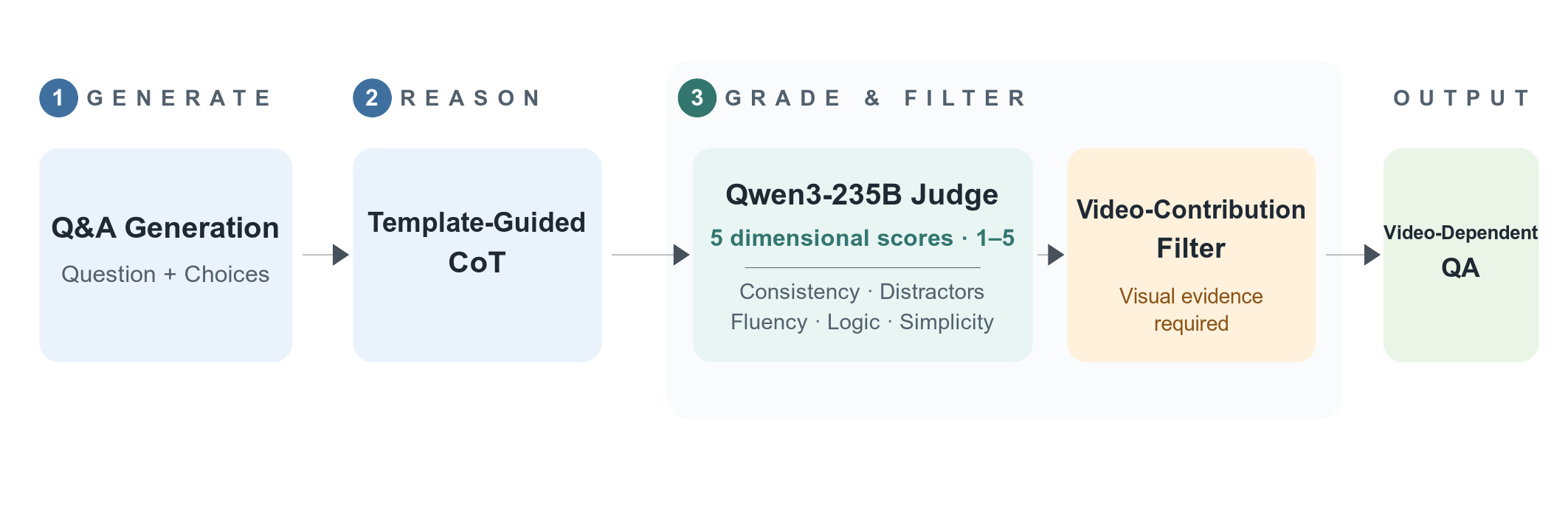}
    \caption{Video-Contribution Filtering for Track 1. Candidates are screened to reduce questions answerable without visual evidence.}
    \label{fig:video-contribution-filtering}
\end{figure}

Following the contribution-aware principle of Audio-Contribution Filtering~\cite{he2025audio}, Video-Contribution Filtering screens candidates for dependence on visual evidence (Figure~\ref{fig:video-contribution-filtering}). This step aims to reduce modality shortcuts~\cite{agrawal2018look} by filtering questions answerable without viewing the video.

Track 2 uses the EgoBench user-agent-tool sandbox~\cite{egobench2026} across retail, kitchen, restaurant, and order scenarios. The final release contains 309 tasks across \texttt{retail6}, \texttt{retail10}, \texttt{kitchen4}, \texttt{restaurant5}, and \texttt{order2}. The \href{https://github.com/ego-link/egolink2026/blob/main/code/track2/EgoBench/README.md}{official final scoring list} identifies a common subset of 96 tasks selected to avoid ground-truth ambiguities: 20 each from \texttt{retail6}, \texttt{retail10}, and \texttt{order2}, and 18 each from \texttt{kitchen4} and \texttt{restaurant5}.

\subsection{Evaluation Metrics}

Track 1 uses overall top-1 MCQ accuracy as the primary metric, with additional scores for emotion, causality, intent, and summarization.

Track 2 ranks systems by \texttt{joint\_success\_rate}: a task achieves joint success only when it passes both tool-call-based process evaluation and database-result-based evaluation~\cite{egobench2026}. Micro accuracy measures call-level tool accuracy, while separate tool-based and result-based success scores diagnose process--result mismatches. Table~\ref{tab:metrics} summarizes these evaluation signals.

\begin{table}[!t]
\centering
\small
\caption{Evaluation signals used in EgoLink.}
\label{tab:metrics}
\begin{tabularx}{\columnwidth}{@{}lY@{}}
\toprule
Metric & Description \\
\midrule
Track 1 accuracy & Overall top-1 MCQ accuracy over the evaluation questions. \\
Dimension scores & Scores for emotion, causality, intent, and summarization question groups. \\
Track 2 joint success & Task-level success requiring both correct tool process and correct database result. \\
Micro accuracy & Call-level tool accuracy for diagnosing fine-grained execution behavior. \\
Result/tool success & Separate result-based and tool-based scores that expose process-result mismatches. \\
\bottomrule
\end{tabularx}
\end{table}

\subsection{Rules}

The EgoLink Challenge follows rules designed to make leaderboard differences reflect reasoning and execution methods rather than hidden data advantages.

\begin{itemize}
    \item For Track 1, participants must pair EgoLink labels with the $E^3$ video data rather than using the original $E^3$ labels.
    \item Manual annotation of the evaluation set is prohibited.
    \item External labeled or unlabeled video data cannot be used for training, fine-tuning, distillation, calibration, or pseudo-label construction in Track 1.
    \item Track 2 submissions must follow the official interaction protocol and output valid natural-language responses or JSON tool-call arrays.
    \item The Track 2 challenge does not allow participants to directly expose the test-set database or user instructions in their prompts.
    \item The challenge materials are released for non-commercial research and education use under the stated license and data-governance policy.
\end{itemize}

\subsection{Participants}

We report official leaderboard scores and analyze four organizer-provided technical reports: GXU-LIPE and Lai for Track 1, and MediaLab and GML-MM Group for Track 2. These reports cover a representative subset of submissions.

\section{Challenge Results}

\begin{table}[t]
\centering
\caption{Official Track 1 leaderboard excerpt.  Accuracy is the overall top-1 MCQ score reported on the challenge website.}
\label{tab:track1leaderboard}
\begin{tabularx}{\columnwidth}{*{3}{>{\centering\arraybackslash}X}}
\toprule
Rank & Team & Accuracy \\
\midrule
1 & GXU-LIPE & 0.9263 \\
2 & Lai & 0.9246 \\
3 & TeleAI & 0.9132 \\
4 & EgoMinds & 0.9092 \\
5 & wjdd & 0.9089 \\
5 & MR-CAS & 0.9089 \\
\bottomrule
\end{tabularx}
\end{table}

\begin{table}[t]
\centering
\caption{Official Track 2 leaderboard excerpt.  Joint success is the primary ranking metric.}
\label{tab:track2leaderboard}
\begin{tabularx}{\columnwidth}{Xcccc}
\toprule
Team & Joint & Micro & Result & Tool \\
\midrule
MediaLab & \textbf{0.844} & \textbf{0.916} & 0.875 & \textbf{0.854} \\
GML-MM Group & 0.823 & 0.914 & 0.854 & 0.823 \\
BigVisionAgentGoGoGo & 0.781 & 0.889 & \textbf{0.896} & 0.802 \\
uestc\_goat & 0.729 & 0.814 & 0.760 & 0.740 \\
EgoMinds & 0.604 & 0.733 & 0.646 & 0.615 \\
\bottomrule
\end{tabularx}
\end{table}

\begin{table*}[t]
\centering
\caption{Representative participant teams and their main technical mechanisms.}
\label{tab:systems}
\begin{tabularx}{\textwidth}{p{0.16\textwidth} p{0.12\textwidth} Y X}
\toprule
Team & Track & Main mechanism & Reported evidence \\
\midrule
GXU-LIPE & Track 1 &
Question-conditioned social evidence graph, candidate-wise verification, source-traceable reranking, low-confidence retraction, and all-option evidence calibration. &
92.89\% validation accuracy and 92.63\% official test accuracy. \\
\midrule
Lai & Track 1 &
Cold-start structured reasoning data, elimination-guided reward design, and option-perturbed DAPO that rewards semantic candidate identity rather than displayed option letters. &
92.75\% validation accuracy and 92.46\% official challenge accuracy. \\
\midrule
MediaLab & Track 2 &
Scenario-aware service agent, official-tool canonicalization, evidence ledger, and scope-limited correction for state-changing calls and final replies. &
0.844 Joint and 0.916 Micro over five test scenarios. \\
\midrule
GML-MM Group & Track 2 &
Status-aware grounded execution, Reporter summaries for evidence compression, step-specific tool views, and paired visual grounding for occluded references. &
0.781 Joint on released \texttt{retail1}; 0.823 Joint over five test scenarios. \\
\bottomrule
\end{tabularx}
\end{table*}

Tables~\ref{tab:track1leaderboard} and~\ref{tab:track2leaderboard} report the official rankings, while Table~\ref{tab:systems} summarizes representative methods.

\subsection{Track 1: Social Reasoning in Egocentric Video}

The leading Track 1 approaches address complementary risks: unsupported social interpretations and sensitivity to answer-option positions.

GXU-LIPE frames answer selection as evidence verification. It builds a question-conditioned social evidence graph and evaluates each candidate for supporting, contradictory, and missing evidence. The system combines these assessments with risk-aware re-verification and all-option evidence calibration to resolve uncertainty among socially plausible candidates.

Lai focuses on the distinction between semantic candidate identity and option position. Its structured cold-start data and EgoAffect reward incorporate question type, emotional tone, and candidate elimination. Option-Perturbed DAPO reorders answer candidates and assigns rewards by semantic identity to reduce reliance on displayed letters. The reported ablation reaches 91.98\% with process-aware reward and 92.75\% validation accuracy with the full option-perturbed setting. These results support further evaluation of candidate discrimination under option permutations.

\subsection{Track 2: Interactive Agent Challenge}

Track 2 highlights the distinction between execution outcomes and tool-process correctness. For example, BigVisionAgentGoGoGo achieves 0.896 result-based success and 0.802 tool-based success. This gap motivates reporting process correctness alongside final-state success.

MediaLab couples a scenario-aware service agent with a correction agent through a shared evidence ledger. Visual hypotheses are canonicalized through official tools before execution, while the correction agent audits high-risk database mutations and final replies against dialogue context, tool observations, and current database state. Removing correction reduces Joint from 0.844 to 0.781; removing scenario rules lowers it to 0.646. The larger drop without scenario rules indicates reliance on scenario-specific guidance in the reported setting.

GML-MM Group emphasizes explicit execution-state tracking and evidence compression. It represents user requests as objectives with dependencies, constraints, and completion status. Reporter summaries retain verified facts, state changes, intermediate computations, and unresolved requirements, while paired grounding across adjacent or cropped frames helps recover occluded object references. This design links progress tracking with visual evidence recovery during multi-step execution.

\section{Conclusion and Future Directions}

EgoLink benchmarks egocentric social reasoning and interactive service-task execution. Across the submissions summarized in Table~\ref{tab:systems}, candidate verification, option-robust reasoning, tool canonicalization, and execution-state tracking highlight the role of explicit intermediate evidence. The reported ablations suggest the value of option-robust training and execution correction in the evaluated settings, while generalization beyond the released scenarios is not established by the results summarized here.

Future work should strengthen evidence grounding and robustness. For Track 1, priorities include modality-dependence tests, option-order perturbations, and temporal evidence evaluation to distinguish grounded reasoning from shortcut-based success~\cite{guo2023desiq,di2024grounded}. For Track 2, evaluation should extend to unseen scenarios, longer interactions, and recoverable tool failures, using step-level progress and database-state checks to assess execution reliability~\cite{lu2025toolsandbox,ma2024agentboard}. These directions aim to strengthen the assessment of auditable reasoning and reliable execution across scenarios.

\bibliographystyle{ACM-Reference-Format}
\balance
\bibliography{references}

\end{document}